\documentclass[10pt,twocolumn,letterpaper]{article}

\usepackage[pagenumbers]{cvpr}

\definecolor{cvprblue}{rgb}{0.21,0.49,0.74}
\usepackage[pagebackref,breaklinks,colorlinks,allcolors=cvprblue]{hyperref}

\def\confName{CVPR}
\def\confYear{2026}

\begin{document}

\title{Stealthy in Semantics, Antagonistic in Space: Attacking Visible-Infrared Object Detectors via Object-Level Misalignment}
\author{%
{\small Yueqi Zhu$^{1,*}$, Qi Ming$^{1,*}$, Guo Cheng$^{1}$, Yongkang Zhang$^{1}$, Feiran Liu$^{1}$, Juan Fang$^{1}$, Jiahuan Zhou$^{2}$, Jiangmeng Li$^{3}$, Yuhan Zhang$^{4}$}\\[-1pt]
{\footnotesize $^{1}$Beijing University of Technology \quad $^{2}$Peking University}\\[-2pt]
{\footnotesize $^{3}$National Key Laboratory of Space Integrated Information System, Institute of Software, Chinese Academy of Sciences}\\[-2pt]
{\footnotesize $^{4}$Intelligent Science \& Technology Academy of CASIC}\\[-2pt]
{\footnotesize $^{*}$Equal contribution. \quad \texttt{15510323848@emails.bjut.edu.cn}}
}
\maketitle

\begin{abstract}

Visible-infrared object detectors are used for robust perception under challenging illumination and weather conditions. Current physical attacks apply conspicuous patches to spatially aligned target regions, which are noticeable to human observers. Meanwhile, most of these methods only perturb the appearance within the aligned region, without explicitly targeting the correspondence between modalities or the fusion process. In this paper, we propose CamoShift, an adversarial framework for visible-infrared object detection. By combining visual camouflage with object-level infrared shifting, CamoShift breaks cross-modal spatial alignment and disrupts fusion. Specifically, the Semantic Camouflage Module (SCM) generates a stealthy camouflaged patch that can be attached to the host object and maintains its effectiveness in the infrared branch through an RGB-IR adapter. The Object-level Spatial Decoupling Module (OSDM) shifts the infrared target evidence in a scale-aware manner, so as to break object-level correspondence and disrupt cross-modal fusion. Then, the Harmonic Adversarial loss (HarAdv loss) further balances attack strength and visual stealth during optimization. To the best of our knowledge, we are the first to target both visual stealthiness and attack success in visible-infrared object detection. Extensive experimental results show that CamoShift achieves a superior balance between attack effectiveness and visual stealth. Code and models will be available on GitHub.

\end{abstract}

\section{Introduction}

\label{sec:intro}
Object detection enables systems to rapidly identify and locate target objects \cite{zou2023object, redmon2016you, karasawa2017multispectral}. Visible-infrared object detectors are increasingly being adopted to tackle challenging adverse environments. While RGB cameras capture fine visual details, infrared sensors provide robust infrared cues that effectively overcome glare and fog \cite{hwang2015multispectral, li2019illumination, sun2022detfusion, dasgupta2022spatio}. Despite their robustness to environmental challenges, the underlying deep learning models remain highly vulnerable to adversarial attacks \cite{evtimov2018robust, thys2019fooling, kim2022defending}.

\begin{figure}[t]
    \centering
    \includegraphics[width=0.98\columnwidth]{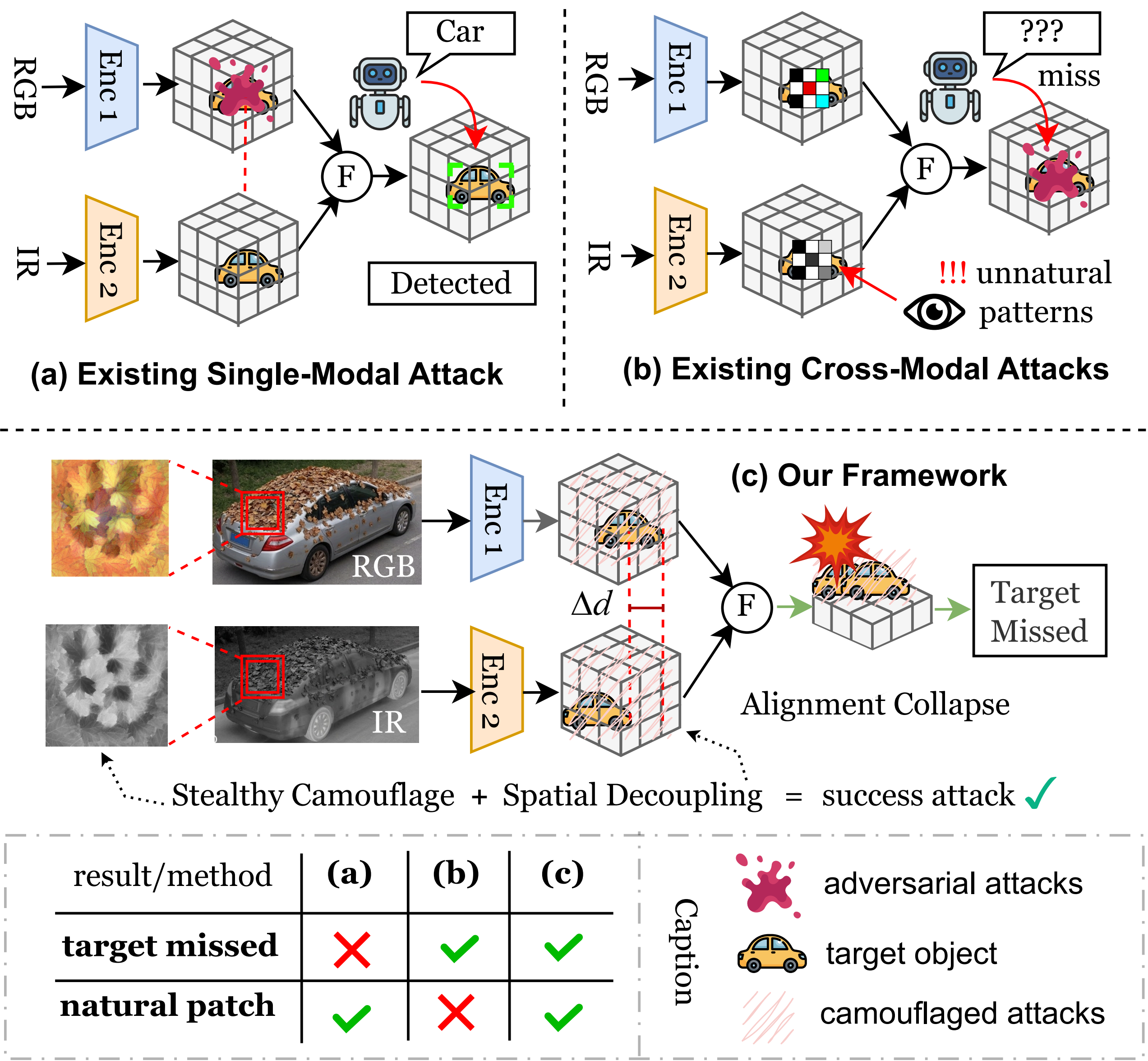}
    \caption{Comparison of adversarial attacks on cross-modal detectors. (a) Single-modal attacks manipulate only visible textures, inherently failing in the infrared domain. (b) Cross-modal adversarial attacks rely on conspicuous artificial patterns and strictly maintain object-level spatial alignment, failing to disrupt the fusion mechanism. (c) Our framework achieves both visual stealthiness and high attack success. It combines semantic camouflage with object-level spatial decoupling to break cross-modal alignment, successfully misleading the detector.}
    \label{fig:motivation}
\end{figure}

Physical adversarial patches aim to mislead object detection systems by deploying crafted patterns or materials in the physical world. To achieve this, mainstream methods have evolved from single-modal to visible-infrared strategies. Early approaches primarily targeted single-modal object detectors by optimizing textures or geometric shapes \cite{brown2017adversarial, wei2022adversarial, xu2020adversarial, hu2022adversarial}. However, these pure RGB patches fail in the infrared domain due to the lack of infrared signatures. To mislead both sensors simultaneously, recent advancements have begun to explore cross-modal physical attacks by deploying crafted multispectral adversarial patches (e.g., combining visual patterns with infrared materials) \cite{wei2024physical, zhu2021fooling, zhu2022infrared}. As a foundational step, CDUPatch \cite{long2025cdupatch} exploits the physical mapping between RGB colors and infrared absorption to generate patches that mislead both branches.

Despite their initial success, existing cross-modal adversarial patch attacks exhibit two critical limitations: \textbf{(1) Highly conspicuous visual appearance.} To generate sufficient infrared differences, existing methods \cite{long2025cdupatch, wei2023unified, zhu2022infrared, hu2025touap} are forced to use extreme RGB color contrasts (e.g., pure black and white blocks). Consequently, these patches manifest as highly unnatural, conspicuous artificial patterns that cannot blend into the background, as shown in Fig.~\ref{fig:motivation}(b). They are instantly detectable by human observers, failing the basic requirement of remaining unnoticeable. \textbf{(2) Strict reliance on object-level spatial alignment.} Current cross-modal detectors fundamentally assume strict spatial alignment between RGB and IR inputs. These systems strictly treat the spatial misalignment as a global, image-level sensor calibration error. Existing attacks \cite{wei2023unified, lee2023multispectral, wei2023hotcold} blindly inherit this assumption, confining their perturbations to corrupting features within perfectly matched bounding boxes. However, they ignore a more fundamental issue that an adversarial patch can actively shift the location of the target object in a single modality. As illustrated in Fig.~\ref{fig:motivation}(c), target features no longer spatially align at the object level across the two modalities, which prevents the network from pairing them and causes the fusion mechanism to fail.

\begin{figure}[t]
    \centering
\includegraphics[width=0.94\columnwidth]{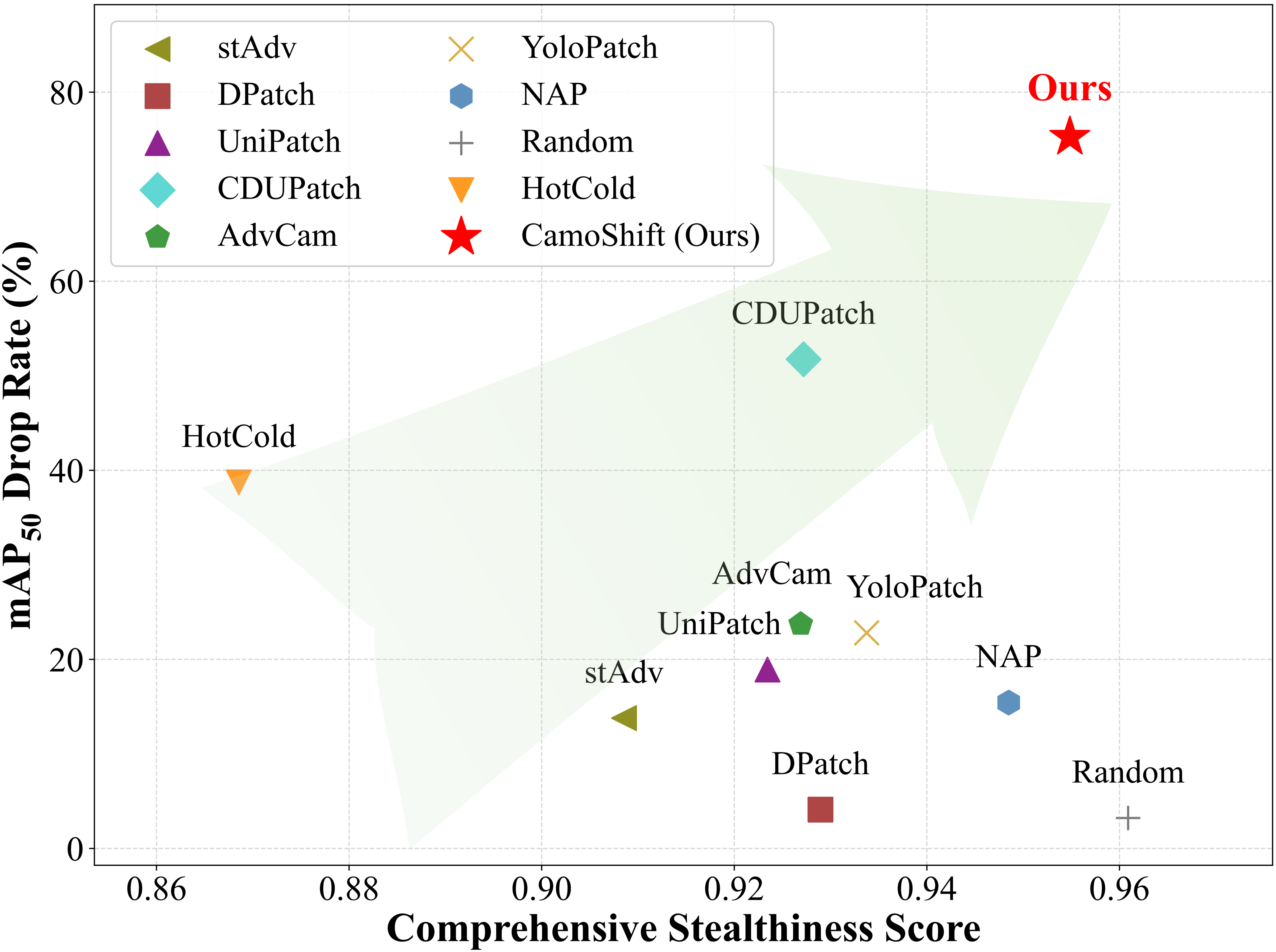}
    \caption{Performance comparison of visual stealthiness and attack success on the DroneVehicle dataset.}
    \label{fig:tradeoff}
\end{figure}

In this paper, to expose the vulnerability of spatial alignment mechanisms and enhance attack stealthiness, we propose a framework named CamoShift. First, we design a Semantic Camouflage Module (SCM) to generate visually stealthy adversarial patches. SCM uses style-consistent, color-driven generation techniques to create adversarial textures, enabling them to seamlessly blend into the background. Compared to traditional attacks that rely on conspicuous color blocks, our method achieves excellent visual deception, thereby reducing the risk of human detection. Second, we introduce an Object-level Spatial Decoupling Module (OSDM) to disrupt cross-modal spatial alignment. OSDM utilizes a scale-aware offset generator to stealthily shift the infrared target at the object level away from its visible location. This manipulation disrupts the strict object-level spatial alignment assumption relied upon by cross-modal detectors. Consequently, this spatial decoupling triggers severe cross-modal feature antagonism. It forces the fusion mechanism to aggregate real features with conflicting background noise, subsequently flattening the network's attention features. Finally, we formulate a Harmonic Adversarial loss (HarAdv loss) to dynamically balance adversarial strength with stealthiness. HarAdv loss allows the attack to adapt robustly to complex environments. As illustrated in Fig.~\ref{fig:tradeoff}, CamoShift achieves SOTA comprehensive performance among existing baselines, which demonstrates that our attack is both effective and visually stealthy.

In summary, our contributions are threefold:
\begin{itemize}
     \item We propose CamoShift to attack visible-infrared object detectors by combining semantic camouflage with object-level spatial decoupling. CamoShift is the pioneering attempt to target both visual stealthiness and attack success in visible-infrared object detection.

    \item We design SCM and OSDM in CamoShift for stealthy cross-modal attacks. Specifically, SCM generates a semantically camouflaged and thermally valid patch, while OSDM spatially shifts the infrared target to break object-level alignment and disrupt fusion.

    \item The HarAdv loss is designed to harmonize attack strength and visual stealth. Experimental results validate that this loss design is crucial to achieving a superior balance between adversarial effectiveness and visual camouflage compared with existing methods.
\end{itemize}

\begin{figure*}[t]
    \centering
    \includegraphics[width=0.94\textwidth]{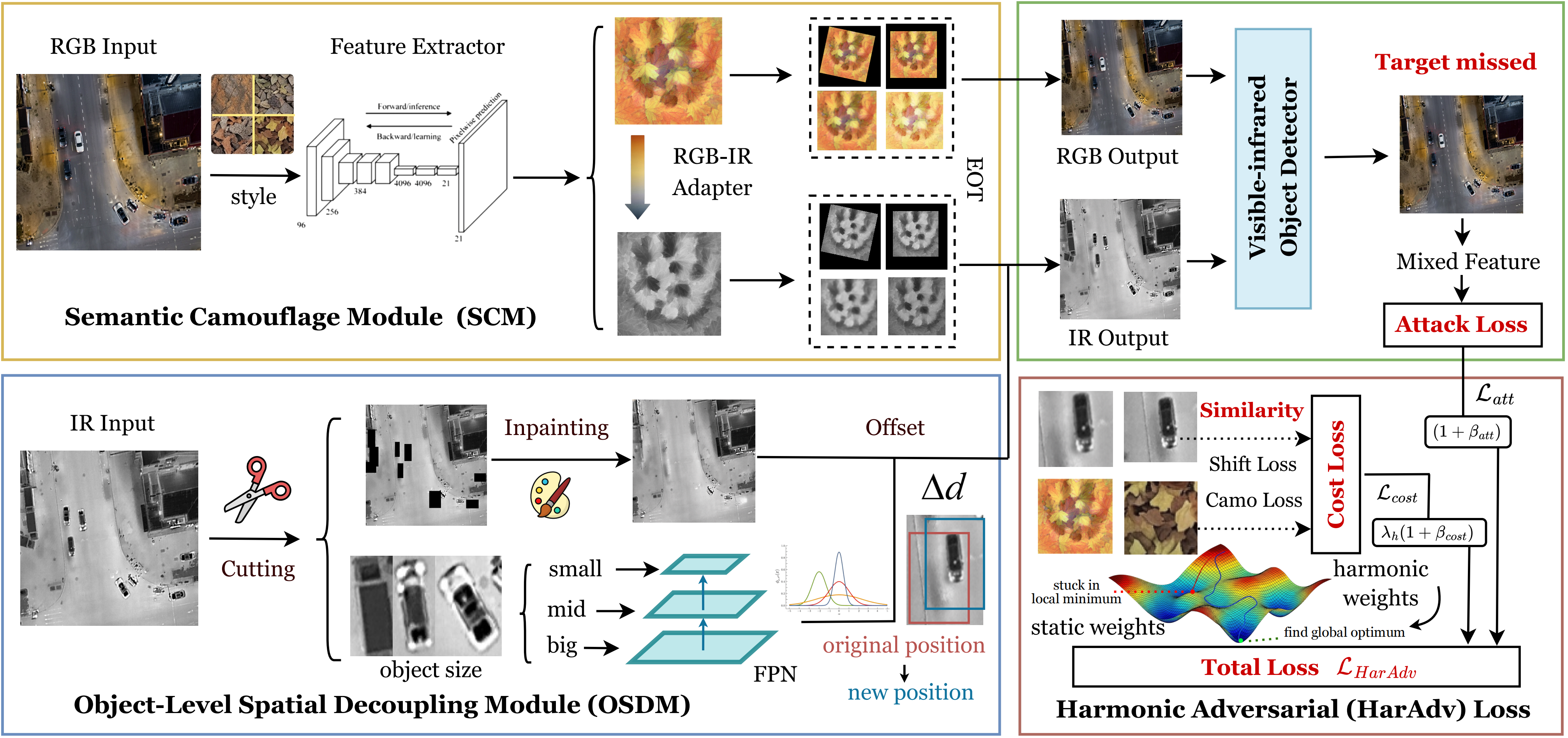}
    \caption{Overview of our CamoShift framework. SCM generates a visually stealthy patch and its valid infrared counterpart via an RGB-IR adapter. Concurrently, OSDM utilizes an FPN to compute a spatial offset for the infrared target, using inpainting and recomposition to break cross-modal spatial alignment. Finally, a harmonic adversarial loss is introduced to dynamically balance the attack effectiveness and visual stealthiness during optimization.}
    \label{fig:overview}
\end{figure*}

\section{Related Work}
\label{sec:related}

\subsection{Visible-Infrared Object Detection}
Visible-infrared object detection combines visual details with infrared features to operate robustly under diverse weather conditions, which is crucial for challenging scenarios like UAV surveillance \cite{sun2022drone, zhang2023superyolo}. While early methods relied on naive feature concatenation \cite{hwang2015multispectral, liu2016multispectral, wagner2016multispectral, xu2017learning}, recent advanced models utilize smarter fusion strategies \cite{cao2022locality, zhang2019cross, he2023multispectral, yuan2024improving, shen2024icafusion, zhang2024when}.

Due to the physical separation of visible and infrared sensors, the images they capture never align perfectly. To mitigate this, modern detectors employ spatial alignment modules. Networks like MBNet \cite{zhou2020mbnet} and C2Former \cite{zhao2023c2former}, along with the offset-guided fusion in COMO \cite{liu2026cross}, and weakly aligned learning methods \cite{lu2021weakly}, calculate dense similarities to explicitly align the features from both sensors. However, this creates a critical vulnerability by assuming the misalignment is merely a global camera shift.

\subsection{Adversarial Attacks}

Early adversarial attacks mainly focused on digital perturbations,
including pixel-level noise and spatial transformations such as
pixel warping \cite{xiao2018spatially}. Their effectiveness often
degrades under physical-world variations. Physical attacks later
introduced printable patches \cite{thys2019fooling,brown2017adversarial}
and wearable materials \cite{wu2020making}. Camouflaged methods such
as AdvCam \cite{duan2020adversarial} and NAP
\cite{hu2021naturalistic} further improve visual stealth, but their
original single-modal setting does not affect both sensor branches.

Cross-modal physical attacks, including UNIPatch
\cite{wei2023unified} and CDUPatch \cite{long2025cdupatch}, use
infrared-blocking materials or visible-to-infrared properties to
attack both modalities. Most methods rely on
high-contrast patterns and confine perturbations to the aligned
target region. CamoShift targets object-level cross-modal
correspondence through semantic camouflage and infrared spatial
decoupling.

\section{Methodology}
\label{sec:method}

\subsection{Overall Structure}

Given an aligned visible-infrared pair $I=(I^{v}, I^{ir})$ and a
target object $o$ with class label $c_o$ and bounding box $b_o$, a
visible-infrared detector $D$ predicts a set of detections $Y=D(I)$.
Current visible-infrared patch attacks mainly perturb the appearance
inside the aligned target box. Although such perturbations may suppress
local confidence, they still preserve the cross-modal correspondence
assumed by the fusion network. We find that, alongside appearance
corruption, the fusion networks of current object detectors are highly
vulnerable to object-level spatial misalignment.

As illustrated in Fig.~\ref{fig:overview}, our CamoShift framework
jointly optimizes a semantically camouflaged visible pattern and a
scale-aware spatial offset in the infrared branch. It consists of three
components: 1) a Semantic Camouflage Module (SCM) that makes the
visible patch blend into the host object while enforcing thermal
feasibility; 2) an Object-level Spatial Decoupling Module (OSDM) that
shifts the infrared evidence to break the alignment prior used by
feature fusion; and 3) a Harmonic Adversarial loss (HarAdv loss) that
balances attack effectiveness and visual stealthiness. For a sampled
transformation, the visible and infrared patching process is written as
follows:
\begin{equation}
\begin{aligned}
&P^{ir} = F_A(P^{v}),\\
&I^{v}_{adv} = (1-M_{\tau})\odot I^{v}
+M_{\tau}\odot \tau(P^{v}),\\
&I^{ir}_{p} = (1-M_{\tau})\odot I^{ir}
+M_{\tau}\odot \tau(P^{ir}),
\end{aligned}
\label{eq:patch_insert_new}
\end{equation}
where $P^{v}$ is the learnable visible patch, $P^{ir}$ is its infrared
counterpart, $I^{v}_{adv}$ and $I^{ir}_{p}$ are the patched visible and
infrared images, $\tau$ is the sampled transformation, $F_A$ is the
RGB-IR adapter, $M_{\tau}$ is the transformed patch mask. $F_A$ learns an empirical
response mapping from paired RGB-IR data rather than reproducing the
thermal imaging process. SCM optimizes the aligned
patched pair, while OSDM moves infrared object evidence away
from the visible target.

\subsection{Semantic Camouflage Module}

SCM is designed to generate visually stealthy adversarial patches. In the visible domain, the patch camouflages itself as the natural surface patterns of the host object while maintaining its attack performance. Meanwhile, in the infrared domain, the same patch must still produce a realistic and effective infrared response through the RGB-IR adapter. Therefore, SCM is not only an appearance regularizer. It is the stage that establishes a valid cross-modal attack state on which the OSDM can operate.

A visually stealthy patch should look like a plausible part of the object surface instead of an external marker \cite{wang2021dual, wei2023moire}. Let $C_o^{v}=\operatorname{Crop}(I^{v}, b_o)$ denote the visible crop of the host object and let $I^{sty}$ be a natural reference texture, such as rust, dried mud, or surface stain. Using a pretrained VGG feature extractor, we define the style, content, and color consistency terms as follows:
\begin{align}
\mathcal{L}_{sty}&=\sum_{l\in S_s}\frac{1}{C_l^2}\left\|G\!\left(\phi_l(P^{v})\right)-G\!\left(\phi_l(I^{sty})\right)\right\|_1, \label{eq:sty_new}\\
\mathcal{L}_{cnt}&=\sum_{l\in S_c}\frac{1}{C_lH_lW_l}\left\|\phi_l(P^{v})-\phi_l(C_o^{v})\right\|_1, \label{eq:cnt_new}\\
\mathcal{L}_{col}&=\left\|\mu(P^{v})-\mu(C_o^{v})\right\|_1+\left\|\sigma(P^{v})-\sigma(C_o^{v})\right\|_1, \label{eq:col_new}
\end{align}
where $\phi_l(\cdot)$ extracts the feature map at the $l$-th layer of VGG, with shape $C_l \times H_l \times W_l$. $S_s$ and $S_c$ denote the sets of layers used for style and content extraction, respectively. $G(\cdot)$ computes the Gram matrix, while $\mu(\cdot)$ and $\sigma(\cdot)$ compute the spatial mean and standard deviation. The style term aligns the patch with the texture family of the reference pattern, the content term keeps the patch anchored to the local structure of the host object, and the color term prevents the patch from degenerating into conspicuous artificial patterns.

Beyond achieving visual stealth, SCM must also effectively suppress the confidence of the detector. To this end, we define an aligned attack loss based on the predictions generated from the patched pair in Eq.~\eqref{eq:patch_insert_new}:
\begin{equation}
\mathcal{L}^{SCM}_{att}=\frac{1}{|Q_o|}\sum_{k\in Q_o}s_k\,p_k(c_o),
\label{eq:latt_scm}
\end{equation}
where $Q_o$ collects the residual predictions associated with the target region, $s_k$ is the objectness score, and $p_k(c_o)$ is the class probability for the target class. Minimizing Eq.~\eqref{eq:latt_scm} makes SCM a true attack module rather than a pure camouflage pre-processor.

The corresponding SCM cost term gathers all appearance-related constraints into a single branch:
\begin{equation}
\begin{aligned}
\mathcal{L}^{SCM}_{cost} &= \lambda_{sty}\mathcal{L}_{sty} + \lambda_{cnt}\mathcal{L}_{cnt} + \lambda_{col}\mathcal{L}_{col} \\
&\quad + \lambda_{tv}\mathcal{L}_{tv} + \lambda_{ssim}\bigl(1-\operatorname{SSIM}(P^{v},C_o^{v})\bigr) \\
&\quad + \lambda_{lpips}\operatorname{LPIPS}(P^{v},C_o^{v}),
\end{aligned}
\label{eq:lcost_scm}
\end{equation}
where $\mathcal{L}_{tv}$ denotes the total variation loss used to enhance local smoothness, SSIM and LPIPS measure structural similarity and perceptual distance, and the $\lambda_{\cdot}$ terms are scalar loss weights. The attack term keeps the patch adversarial, while the cost term keeps it semantically camouflaged and visually printable. All SCM terms are optimized under Expectation Over Transformation (EOT), so the patch is not tied to a fixed placement or orientation.

\subsection{Object-level Spatial Decoupling Module}
OSDM disrupts object-level alignment by shifting target
evidence only in the infrared branch. The modeled shift represents
RGB-IR mismatch in moving scenes caused by differences in exposure
time, frame rate, or sensor readout. Coupled with SCM, the resulting
mismatch breaks the alignment prior used by cross-modal fusion.

To ensure the offset is both effective and stealthy, we utilize a feature pyramid network (FPN) to adaptively adjust the offset based on the object scale. For each target object, the pyramid level is assigned according to the object size:
\begin{equation}
l_{sel}=\arg\min_{l}\left|\log_2\frac{\sqrt{w_oh_o}}{s_l}-\kappa\right|, \label{eq:level_new}
\end{equation}
where $w_o$ and $h_o$ are the width and height of the target box, $l$ indexes the pyramid levels, $s_l\in\{8,16,32\}$ is the stride of level $l$, and $\kappa$ is the canonical scale constant. We then extract the corresponding RoI feature $z_o$ and predict a bounded offset via a generator network $G_{\theta}$:
\begin{equation}
\Delta d_o=(\Delta x_o,\Delta y_o)=\alpha s_{l_{sel}}\tanh\bigl(G_{\theta}(z_o)\bigr),
\label{eq:offset_new}
\end{equation}
where $\Delta x_o$ and $\Delta y_o$ are the horizontal and vertical offsets, $\theta$ denotes the parameters of $G_{\theta}$, and $\alpha$ is set to 1.5 in our experiments. The hyperbolic tangent bounds the displacement, while the factor $\alpha s_{l_{sel}}$ ties the offset magnitude to the receptive-field stride of the assigned pyramid level. As a result, the infrared target is encouraged to cross the corresponding grid-cell boundary on the feature map, even when the image-plane shift remains visually small.

However, directly translating the infrared target creates empty holes and sharp boundaries in the background. To synthesize a plausible misaligned pair, we first remove the original target from the infrared image by inpainting, and then paste the shifted target back with a soft mask:
\begin{equation}
\begin{aligned}
&\hat{I}^{ir}=H(I_p^{ir},M_o),\\
&I^{ir}_{adv}=\bigl(1-S_{\Delta d_o}(M_o^{s})\bigr)\odot \hat{I}^{ir}+S_{\Delta d_o}\!\left(M_o^{s}\odot I_p^{ir}\right),
\end{aligned}
\label{eq:recompose_new}
\end{equation}
where $\hat{I}^{ir}$ is the inpainted infrared image, $I^{ir}_{adv}$ is the shifted infrared image, $M_o$ is the target mask, $M_o^{s}$ is its Gaussian-smoothed version, $H(\cdot)$ denotes LaMa-based inpainting, and $S_{\Delta d_o}(\cdot)$ is the spatial translation operator.

After Eq.~\eqref{eq:recompose_new}, the visible branch still observes the target at the original position, whereas the infrared branch places the main infrared evidence at the shifted position. This spatial mismatch breaks the strict alignment prior relied upon by cross-modal fusion, inducing the cross-modal feature antagonism discussed in the introduction. To describe this mechanism more clearly, we separate the OSDM attack term from the OSDM cost term.

The OSDM attack branch has three parts. The first reduces the residual confidence around the original and shifted semantic positions. The second weakens visible-infrared agreement at the original target region. The third flattens the fused response over the two competing regions:
\begin{equation}
\begin{array}{@{}l@{\;}c@{\;}l@{}}
\mathcal{L}_{dis} & = & \displaystyle \frac{1}{|R_o|}\sum_{u\in R_o}
\frac{\langle \Phi_u^{v},\Phi_u^{ir}\rangle}
{\|\Phi_u^{v}\|_2\|\Phi_u^{ir}\|_2+\varepsilon},\\[0.4ex]
\mathcal{L}_{flat} & = & \displaystyle \operatorname{Std}\!\left(\Phi^{fus}\big|_{R_o\cup R_o^{\Delta}}\right),
\end{array}
\label{eq:dis_flat_new}
\end{equation}
where $R_o$ and $R_o^{\Delta}$ denote the projected regions of the original box and the shifted box on the selected feature map. $\Phi_u^{v}$ and $\Phi_u^{ir}$ represent the visible and infrared feature vectors at spatial location $u$. $\Phi^{fus}$ is the fused feature representation, $\Phi^{fus}|_R$ denotes the fused features restricted to region $R$, $\operatorname{Std}(\cdot)$ denotes standard deviation, and $\varepsilon$ is a small constant for numerical stability. The similarity term measures how strongly the two modalities match at the original location, and the flattening term captures whether the fused representation still forms a concentrated target response.

We further define the OSDM detection-suppression term and the full OSDM attack loss as follows:
\begin{equation}
\begin{array}{@{}l@{\;}c@{\;}l@{}}
\mathcal{L}^{OSDM}_{det} & = & \displaystyle \frac{1}{|Q_o^{\Delta}|}\sum_{k\in Q_o^{\Delta}}s_k\,p_k(c_o),\\[0.4ex]
\mathcal{L}^{OSDM}_{att} & = & \displaystyle \lambda_{det}\mathcal{L}^{OSDM}_{det}
+\lambda_{dis}\mathcal{L}_{dis}+\lambda_{flat}\mathcal{L}_{flat},
\end{array}
\label{eq:latt_osdm}
\end{equation}
where $Q_o^{\Delta}$ collects the predictions overlapping either the original visible position or the shifted infrared position, and $\lambda_{det}$, $\lambda_{dis}$, and $\lambda_{flat}$ are scalar loss weights. Eq.~\eqref{eq:latt_osdm} makes OSDM explicit: it is not merely moving pixels, but jointly suppressing prediction confidence, cross-modal agreement, and fused localization concentration after the misalignment is introduced. The OSDM cost term is defined as:
\begin{equation}
\mathcal{L}^{OSDM}_{cost}=\lambda_{reg}\frac{\|\Delta d_o\|_1}{s_{l_{sel}}},
\label{eq:lcost_osdm}
\end{equation}
where $\lambda_{reg}$ is a scalar loss weight. This term penalizes large offsets. Therefore, OSDM seeks the smallest object-level shift needed to break the fusion prior.

\subsection{Harmonic Adversarial Loss}

An attack should remain effective without sacrificing visual stealthiness. Based on the above decomposition, the final attack loss and the final cost loss are no longer written in a mixed and unordered manner. Instead, they are assembled module by module:
\begin{equation}
\begin{array}{@{}l@{\;}c@{\;}l@{}}
\mathcal{L}_{att} & = & \displaystyle \omega_{s}\mathcal{L}^{SCM}_{att}+\omega_{o}\mathcal{L}^{OSDM}_{att},\\[0.4ex]
\mathcal{L}_{cost} & = & \displaystyle \rho_{s}\mathcal{L}^{SCM}_{cost}+\rho_{o}\mathcal{L}^{OSDM}_{cost},
\end{array}
\label{eq:global_attack_cost}
\end{equation}
where $\omega_s$ and $\omega_o$ weight the SCM and OSDM
attack losses, $\rho_s$ and $\rho_o$ weight their
corresponding cost losses. SCM contributes the aligned adversarial pressure and semantic camouflage cost, while OSDM contributes the misalignment-driven adversarial pressure and the offset regularization cost. This makes the source of each optimization signal transparent.

A fixed linear combination of the final attack loss and the cost loss is often unstable because the desired optimization balance shifts during training \cite{kendall2018multi}. We therefore use cross-adaptive harmonic weights:
\begin{equation}
\beta_{att}=\exp\!\left(-\frac{\operatorname{sg}(\mathcal{L}_{cost})}{T_{cost}}\right),
\beta_{cost}=\exp\!\left(-\frac{\operatorname{sg}(\mathcal{L}_{att})}{T_{att}}\right),
\label{eq:beta_new}
\end{equation}
where $\operatorname{sg}(\cdot)$ is the stop-gradient operator
and $T_{att},T_{cost}$ are temperature parameters. The final HarAdv loss is then defined
as follows:
\begin{equation}
\mathcal{L}_{HarAdv}=(1+\beta_{att})\mathcal{L}_{att}+\lambda_h(1+\beta_{cost})\mathcal{L}_{cost},
\label{eq:haradv_new}
\end{equation}
where $\lambda_h$ controls the contribution of the cost loss. When the attack becomes strong but visually less plausible, the harmonic weight of the attack branch decreases and the optimization focuses on camouflage. Conversely, when the patch is plausible but the detector remains stable, the cost branch is relatively down-weighted and the optimization shifts toward attack effectiveness.

\begin{table*}[t]
\centering
\small
\setlength{\tabcolsep}{0.95pt}
\renewcommand{\arraystretch}{1.3}
\resizebox{\textwidth}{!}{%
\begin{tabular}{@{}l*{12}{c}@{}}
\toprule
& \multicolumn{3}{c}{DroneVehicle~\cite{sun2022drone}}
& \multicolumn{3}{c}{M3FD~\cite{liu2022target}}
& \multicolumn{3}{c}{LLVIP~\cite{jia2021llvip}}
& \multicolumn{3}{c}{VEDAI~\cite{razakarivony2016vehicle}} \\
\cmidrule(lr){2-4} \cmidrule(lr){5-7} \cmidrule(lr){8-10} \cmidrule(lr){11-13}
Method & YOLOv8 & C2Former & COMO & YOLOv8 & C2Former & COMO & YOLOv8 & C2Former & COMO & YOLOv8 & C2Former & COMO \\
\midrule
Clean & 84.09 & 91.03 & 88.14 & 77.02 & 70.37 & 85.78
      & 92.47 & 89.23 & 92.80 & 88.86 & 84.63 & 86.27 \\
\midrule
RandomPatch$^\dagger$~\cite{brown2017adversarial} & 4.76/80.07 & 5.39/87.66 & 3.28/85.30 & 2.02/76.53 & 2.21/68.54 & 2.88/84.71 & 4.02/90.76 & 3.76/87.20 & 7.22/90.51 & 17.56/79.56 & 13.36/78.18 & 9.38/74.27 \\
DPatch$^\dagger$~\cite{liu2018dpatch} & 5.29/80.62 & 5.85/86.63 & 5.34/84.52 & 8.04/75.59 & 8.12/68.62 & 7.77/84.55 & 9.33/90.35 & 9.51/86.51 & 9.42/90.77 & 25.78/67.27 & 24.05/66.23 & 28.39/59.73 \\
stAdv$^\dagger$~\cite{xiao2018spatially} & 46.92/65.98 & 27.85/83.91 & 29.82/75.99 & 44.09/56.70 & 46.14/47.97 & 34.55/70.43 & 25.85/80.97 & 18.03/75.87 & 10.65/89.79 & 54.44/59.32 & 55.95/53.55 & 55.40/59.63 \\
AdvCam$^\dagger$~\cite{duan2020adversarial} & 43.39/70.23 & 60.15/70.05 & 53.74/67.22 & 61.02/46.39 & 38.92/53.11 & 41.23/64.23 & 29.67/79.75 & 15.68/77.23 & 32.70/80.35 & 40.67/75.88 & 30.92/71.91 & 56.40/73.99 \\
NAP$^\dagger$~\cite{hu2021naturalistic} & 29.24/76.19 & 39.70/79.56 & 37.23/74.54 & 40.67/59.24 & 25.61/64.12 & 31.62/70.96 & 30.02/81.60 & 14.82/79.42 & 34.26/86.02 & 55.11/69.04 & 27.48/73.22 & 49.29/69.21 \\
YOLOPatch$^\dagger$~\cite{thys2019fooling} & 36.34/73.60 & 46.72/78.17 & 51.90/68.04 & 49.40/50.34 & 35.15/52.99 & 38.18/65.89 & 20.14/85.93 & 14.39/77.88 & 20.35/86.80 & 46.67/78.68 & 24.43/75.97 & 35.07/78.97 \\
FCA$^\dagger$~\cite{wang2022fca} & 16.88/80.93 & 48.85/77.52 & 46.95/70.47 & 40.22/58.35 & 22.93/63.64 & 26.37/75.02 & 31.54/78.32 & 11.97/82.77 & 37.67/76.67 & 34.67/77.27 & 17.94/75.98 & 42.65/73.34 \\
HOTCOLD$^\dagger$~\cite{wei2023hotcold} & 47.21/61.54 & 59.37/58.78 & 62.70/54.01 & 28.95/71.28 & 36.19/55.49 & 29.79/76.00 & 24.07/84.13 & 20.95/73.22 & 35.38/78.54 & 37.78/83.37 & 17.56/79.54 & 38.39/78.23 \\
UNIPatch~\cite{wei2023unified} & 51.20/62.98 & 41.08/75.13 & 39.06/71.46 & 40.42/59.99 & 32.56/57.58 & 51.16/59.81 & 23.63/80.45 & 16.26/79.65 & 22.84/82.03 & 54.22/69.37 & 39.69/66.50 & 53.51/69.64 \\
TOUAP~\cite{hu2025touap} & 34.07/72.64 & 48.99/72.97 & 48.02/68.19 & 38.13/59.89 & 44.91/34.37 & 44.28/58.47 & 20.40/82.57 & 20.08/67.72 & 20.09/84.62 & 45.33/74.95 & 20.99/73.02 & 51.66/65.99 \\
CDUPatch~\cite{long2025cdupatch} & 51.64/60.52 & 56.60/68.90 & 54.54/42.53 & 46.77/58.37 & 47.00/39.39 & 44.40/60.59 & 18.07/86.97 & 31.34/59.33 & 24.29/82.24 & 49.33/66.81 & 26.34/72.09 & 49.76/64.22 \\
\midrule
\textbf{CamoShift (ours)} & \textbf{78.44/20.26} & \textbf{86.96/19.40} & \textbf{64.23/21.81} & \textbf{76.71/27.52} & \textbf{67.08/15.95} & \textbf{65.84/28.71} & \textbf{64.28/25.27} & \textbf{73.64/22.43} & \textbf{57.32/25.62} & \textbf{67.72/27.52} & \textbf{71.72/39.91} & \textbf{62.50/33.45} \\
\bottomrule
\end{tabular}%
}
\caption{Overall comparison on four visible-infrared datasets. The Clean row reports mAP$_{50}$ (\%). All attack rows report
Strict-ASR (\%) / post-attack mAP$_{50}$ (\%). Methods marked
with $\dagger$ are single-modal attacks adapted to the
visible-infrared setting. }
\label{tab:main_results}
\end{table*}

To demonstrate the necessity of the stop-gradient operator, we analyze the gradient of the total objective with respect to the learnable parameters. Let the learnable parameters $W$ contain the visible patch and the OSDM offset generator. Since the dynamic weights are detached from the opposite branch, the gradient simplifies to:
\begin{equation}
\begin{aligned}
&\frac{\partial \mathcal{L}_{HarAdv}}{\partial W} = (1+\beta_{att})\frac{\partial \mathcal{L}_{att}}{\partial W} + \lambda_h(1+\beta_{cost})\frac{\partial \mathcal{L}_{cost}}{\partial W}, \\[1.5ex]
&\frac{\partial}{\partial W}\Bigl((1+\beta_{att})\mathcal{L}_{att}\Bigr) = (1+\beta_{att})\frac{\partial \mathcal{L}_{att}}{\partial W} + \mathcal{L}_{att}\frac{\partial \beta_{att}}{\partial W}.
\end{aligned}
\label{eq:grad_new}
\end{equation}
The first line is the gradient actually used by our optimization. The second line shows the pathological leakage that would appear without stop-gradient. Because $\beta_{att}$ is inversely controlled by the cost branch, the extra term would encourage the optimizer to increase the cost merely to shrink the attack weight, thereby creating a harmful shortcut that ruins camouflage. The stop-gradient prevents this leakage and ensures that the two branches only interact through scalar balancing rather than gradient contamination.

The final parameters are obtained by minimizing the expectation of the HarAdv loss under the transformation distribution, and the optimized infrared patch is generated by the RGB-IR adapter from the optimized visible patch.

\section{Experiments}
\label{sec:experiments}

\subsection{Experimental Setup}
\label{sec:exp_setup}

We conduct experiments on four widely used visible-infrared datasets: DroneVehicle \cite{sun2022drone}, M3FD \cite{liu2022target}, LLVIP \cite{jia2021llvip}, and VEDAI \cite{razakarivony2016vehicle}. We select YOLOv8 \cite{jocher2023yolo}, C2Former \cite{zhao2023c2former}, and COMO \cite{liu2026cross} as our victim models. While YOLOv8 serves as a strong baseline, C2Former and COMO explicitly model cross-modal interaction, providing a strict protocol to evaluate whether our attack effectively disrupts the fusion correspondence prior.

We compare generic attacks, including RandomPatch
\cite{brown2017adversarial}, DPatch \cite{liu2018dpatch},
YOLOPatch \cite{thys2019fooling}, and stAdv
\cite{xiao2018spatially}. Other baselines include the camouflage
attacks AdvCam \cite{duan2020adversarial}, NAP
\cite{hu2021naturalistic}, and FCA \cite{wang2022fca}, the
infrared attack HOTCOLD \cite{wei2023hotcold}, and the
visible-infrared attacks UNIPatch \cite{wei2023unified}, TOUAP
\cite{hu2025touap}, and CDUPatch \cite{long2025cdupatch}.
Single-modal baselines are applied to visible images, while the same RGB-IR adapter used by CamoShift generates their infrared perturbations. Thus, all methods are evaluated under a unified cross-modal protocol.

We introduce Strict Attack Success Rate (Strict-ASR) to require suppression at both semantic locations. Let $N$ be the number of samples, $Y_i(b)$ the predictions
matched to box $b$, $b_o^{(i)}$ the original box,
$b_o^{\Delta(i)}$ the shifted box, and $\mathbb{I}(\cdot)$
the indicator function:
\begin{equation}
\text{Strict-ASR}=\frac{1}{N}\sum_{i=1}^{N}\mathbb{I}\!\left(|Y_i(b_o^{(i)})|=0\land|Y_i(b_o^{\Delta(i)})|=0\right).
\label{eq:strict_asr}
\end{equation}
For baselines without spatial shifting, we set
$b_o^{\Delta(i)}=b_o^{(i)}$. Therefore,
Eq.~\eqref{eq:strict_asr} is mathematically equivalent to
conventional ASR for these baselines. On DroneVehicle with YOLOv8, CamoShift achieves a
conventional ASR of 93.7\%, while its Strict-ASR is 78.4\%. We also report post-attack $\mathrm{mAP}_{50}$, SSIM, and LPIPS.

For Table~\ref{tab:main_results}, patches are optimized on the
corresponding COMO model for each dataset and evaluated on
YOLOv8, C2Former, and COMO. All detectors are trained on their respective datasets, with
clean mAP$_{50}$ reported in Table~\ref{tab:main_results}.
CamoShift is implemented in PyTorch and optimized with Adam for
50 epochs. Implementation details, including hyperparameters,
runtime, memory usage, and hardware, are provided in the
Supplementary Material.

\subsection{Main Results}
\label{sec:main_results}

Table~\ref{tab:main_results} summarizes the main results on all datasets and detectors. A consistent pattern can be observed across all settings. Methods that mainly perturb appearance, including generic baselines and RGB-specific camouflage attacks, have a limited impact in the visible-infrared setting. Their perturbation may reduce the confidence of one branch, but the complementary modality still preserves enough evidence for the detector to localize the target. This effect is particularly clear on COMO, where the retained $\text{mAP}_{50}$ of AdvCam and NAP remains high across all datasets.

Recent cross-modal attacks achieve much stronger results, but their gains remain limited when the perturbation is still constrained to the aligned object region. For example, CDUPatch clearly outperforms single-modality baselines, yet its Strict-ASR on COMO still only ranges from 24.29\% to 54.54\% across the four datasets. At the same time, the detector still retains substantial performance after the attack. This suggests that appearance corruption alone is not sufficient once the detector can continue to associate visible and infrared evidence at the same object position.

\begin{figure}[!t]
    \centering
    \includegraphics[width=0.96\columnwidth]{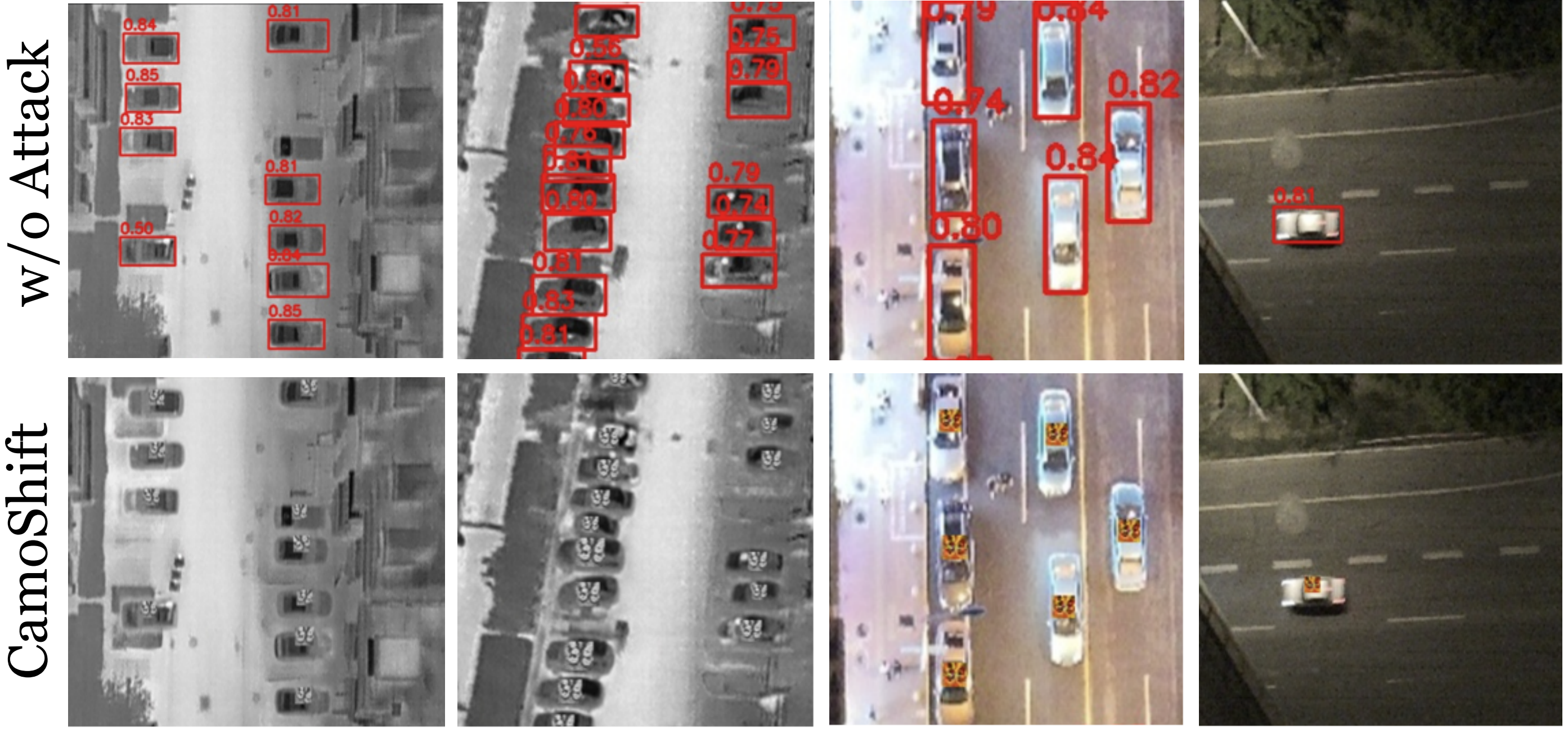}
    \caption{Qualitative results of CamoShift against the COMO detector. To better visualize the attack effects, visible-light images are utilized as the base for daylight scenarios, whereas shifted infrared images are employed for dark environments.}
    \label{fig:qualitative_como}
\end{figure}

In contrast, CamoShift achieves the best results in all settings. On COMO, our method raises Strict-ASR to 64.23\%, 65.84\%, 57.32\%, and 62.50\% on the four datasets, while reducing the retained $\text{mAP}_{50}$ to 21.81\%, 28.71\%, 25.62\%, and 33.45\%, respectively. Similar trends are also observed on YOLOv8 and C2Former. The qualitative attack effects on the COMO detector are further visualized in Fig.~\ref{fig:qualitative_como}. These results support our main claim: once the infrared features are spatially shifted from the visible target, the fusion module is forced to aggregate mismatched cues, leading to a substantially more severe performance degradation than that caused by spatially aligned perturbations.

\subsection{Component-wise Ablation Study}
\label{sec:ablation}

Table~\ref{tab:ablation} presents the ablation study on DroneVehicle using COMO. The RGB-IR adapter alone already produces a valid visible-infrared patch, but its visual quality is poor, with only 0.892 SSIM. Adding semantic camouflage greatly improves the visual appearance to 0.937 SSIM, although the attack success rate drops from 53.1\% to 44.2\%. Stronger camouflage constrains the optimization space, making effective attacks harder to generate.

\begin{table}[!t]
\centering
\small
\setlength{\tabcolsep}{2.5pt}
\renewcommand{\arraystretch}{1.0}
\begin{tabular*}{\columnwidth}{@{\extracolsep{\fill}}cccccc@{}}
\toprule
IR-Ada & Camo & Spat & HarAdv & Strict-ASR $\uparrow$ & SSIM $\uparrow$ \\
\midrule
           &            &            &            & 32.3          & 0.972 \\
\checkmark &            &            &            & 53.1          & 0.892 \\
\checkmark & \checkmark &            &            & 44.2          & 0.937 \\
\checkmark & \checkmark & \checkmark &            & 60.2          & 0.923 \\
\checkmark & \checkmark & \checkmark & \checkmark
           & \textbf{64.2} & \textbf{0.975} \\
\bottomrule
\end{tabular*}
\caption{Ablation study. Strict-ASR is reported in percent.
IR-Ada denotes the RGB-IR adapter; Camo, semantic camouflage; Spat, OSDM; and HarAdv, the harmonic adversarial loss.}
\label{tab:ablation}
\end{table}

The key gain appears when a spatial shift is introduced. Once OSDM is enabled, the Strict-ASR rises sharply to 60.2\% while the patch still preserves reasonable visual quality. This confirms that the main contribution of CamoShift is not a stronger texture perturbation, but the deliberate breaking of object-level cross-modal correspondence. After adding the harmonic optimization objective, both attack strength and stealth improve further, reaching 64.2\% Strict-ASR and 0.975 SSIM. This result indicates that HarAdv loss is effective in balancing the conflict between attack loss and camouflage constraints during optimization.

\subsection{Stealthiness Evaluation}
\label{sec:stealthiness}
Attack strength alone is not sufficient for a patch, since an obviously artificial pattern is easy to detect and remove in practice. We therefore compare the visual quality of the patched region against the original host appearance. The results in Table~\ref{tab:stealthiness} show a clear advantage of the proposed semantic camouflage module. Compared with CDUPatch, our patch improves SSIM from 0.9464 to 0.9745 and reduces LPIPS from 0.0920 to 0.0648 on DroneVehicle. More importantly, this gain is achieved in the visible-infrared setting, where the patch must remain visually plausible while still inducing an effective infrared response.

\begin{table}[!t]
\centering
\small
\renewcommand{\arraystretch}{1.0}
\begin{tabular*}{0.84\columnwidth}{
    @{\hspace{1.4em}}
    l
    @{\extracolsep{\fill}}
    c
    c
    @{\hspace{1.4em}}
}
\toprule
\multicolumn{1}{c}{Method} & SSIM $\uparrow$ & LPIPS $\downarrow$ \\
\midrule
HOTCOLD~\cite{wei2023hotcold}
    & 0.8855 & 0.1484 \\
DPatch~\cite{liu2018dpatch}
    & 0.9303 & 0.0724 \\
UNIPatch~\cite{wei2023unified}
    & 0.9390 & 0.0921 \\
CDUPatch~\cite{long2025cdupatch}
    & 0.9464 & 0.0920 \\
AdvCam~\cite{duan2020adversarial}
    & 0.9476 & 0.0938 \\
YOLOPatch~\cite{thys2019fooling}
    & 0.9483 & 0.0808 \\
\textbf{CamoShift (ours)}
    & \textbf{0.9745}
    & \textbf{0.0648} \\
\bottomrule
\end{tabular*}
\caption{Stealthiness comparison on DroneVehicle using COMO. SSIM and LPIPS compare the patched and original regions.}
\label{tab:stealthiness}
\end{table}

The result is also competitive with single-modal camouflage attacks. AdvCam produces a natural visible pattern, but it does not need to satisfy the cross-modal constraint. In contrast, CamoShift preserves or even improves visible stealth while remaining effective against both modalities. This behavior is consistent with the design of SCM: the style, content, and color regularizers keep the patch close to the host texture distribution, while the RGB-IR adapter ensures that the optimized appearance remains meaningful in the infrared branch.

\begin{figure}[!t]
    \centering
    \includegraphics[width=\linewidth]{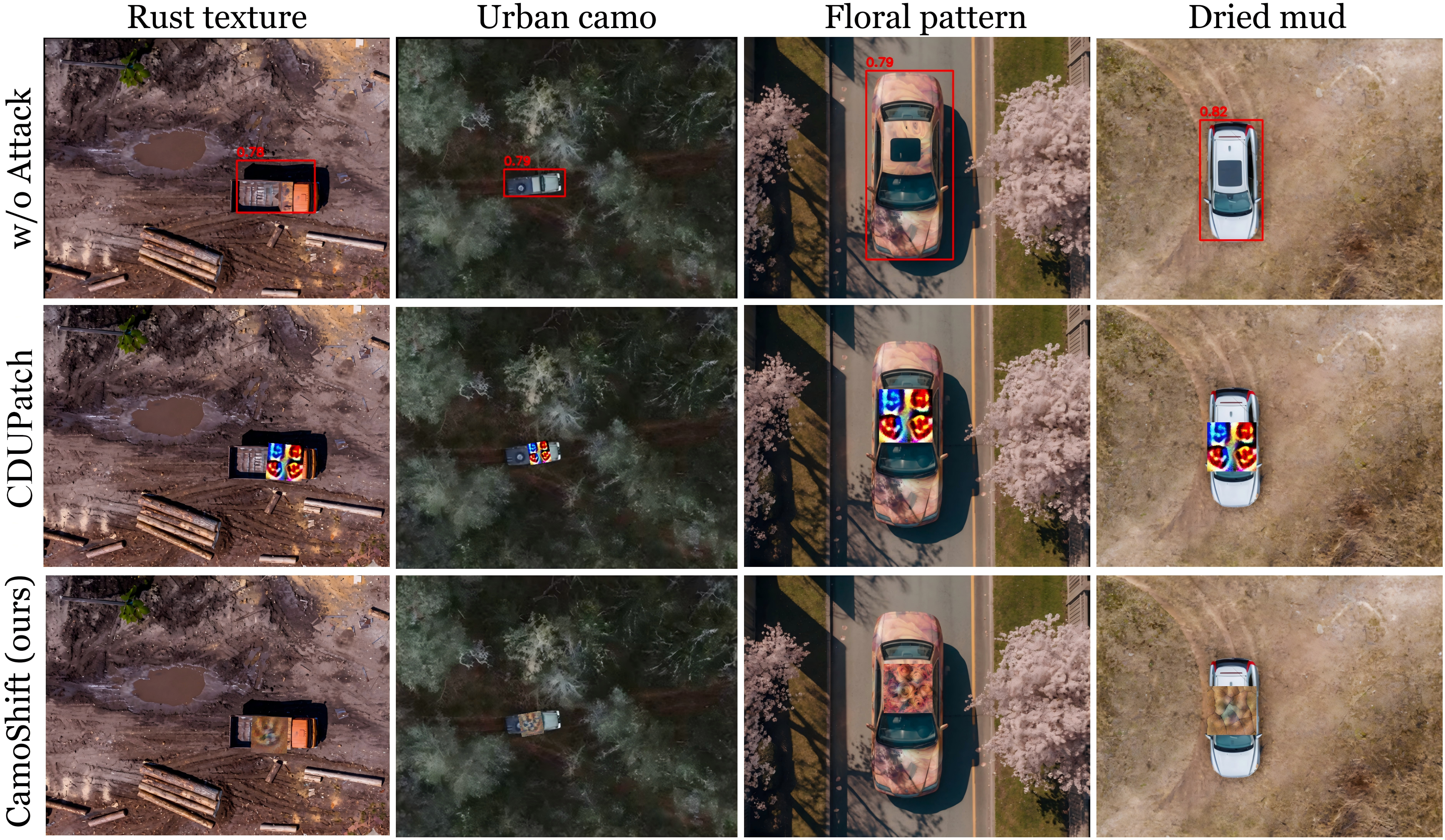}
    \caption{Qualitative results of CDUPatch and our CamoShift across diverse environments.}
    \label{fig:camouflage_styles}
\end{figure}

We also test whether the attack depends on a specific visible appearance. As visualized in Fig.~\ref{fig:camouflage_styles}, Table~\ref{tab:style_robustness} shows that different natural textures, including rust, dried mud, and urban camouflage, all lead to high attack success while preserving good visual similarity. The gap between these styles is small compared with the gap between our method and CDUPatch. This observation is important for deployment, since the patch can be adapted to different environments without changing the underlying attack mechanism.

\begin{table}[!t]
\centering
\small
\setlength{\tabcolsep}{4pt}
\renewcommand{\arraystretch}{1.1}
\begin{tabular}{lcccc}
\toprule
Style & Strict-ASR $\uparrow$ & mAP$_{50}$ $\downarrow$
& SSIM $\uparrow$ & LPIPS $\downarrow$ \\
\midrule
Rust texture   & \textbf{66.21} & 22.57          & \textbf{0.9738} & 0.0602 \\
Dried mud      & 65.42          & 22.62          & 0.9735          & \textbf{0.0592} \\
Urban camo     & 65.28          & \textbf{22.51} & 0.9732          & 0.0603 \\
Floral pattern & 66.19          & 22.56          & 0.9734          & 0.0610 \\
\bottomrule
\end{tabular}
\caption{Effect of different camouflage styles on COMO.
Strict-ASR and mAP$_{50}$ are reported in percent.}
\label{tab:style_robustness}
\end{table}

\subsection{Generalization and Robustness}
\label{sec:robustness}

We further evaluate whether the proposed attack generalizes beyond the white-box setting. Table~\ref{tab:transferability} reports cross-model transfer results on DroneVehicle. A patch optimized on YOLOv8 still reaches 90.61\% Strict-ASR on C2Former and 59.26\% on COMO. Similarly, the patch trained on COMO transfers to YOLOv8 and C2Former with 78.44\% and 86.96\% Strict-ASR, respectively. This result indicates that the gain of CamoShift does not mainly come from overfitting to a specific detector head. Instead, it exposes a common vulnerability in visible-infrared object detectors by exploiting their heavy reliance on consistent \mbox{cross-modal correspondence}.

\begin{table}[!t]
\centering
\small
\setlength{\tabcolsep}{3.5pt}
\renewcommand{\arraystretch}{1.}
\begin{tabular}{lccc}
\toprule
Source model & YOLOv8~\cite{jocher2023yolo} & C2Former~\cite{zhao2023c2former} & COMO~\cite{liu2026cross} \\
\midrule
YOLOv8~\cite{jocher2023yolo}   & \textbf{88.86} & 90.61          & 59.26 \\
C2Former~\cite{zhao2023c2former} & 82.78          & \textbf{91.26} & 63.95 \\
COMO~\cite{liu2026cross}     & 78.44          & 86.96          & \textbf{64.23} \\
\bottomrule
\end{tabular}
\caption{Evaluation of cross-model adversarial transferability.
Rows denote source models and columns denote target models.
Results are evaluated by Strict-ASR (\%).}
\label{tab:transferability}
\end{table}

We further evaluate CamoShift against three common defenses on
DroneVehicle with YOLOv8. Strict-ASR remains 52.1\%, 73.3\%,
and 74.5\% under adversarial training, feature denoising, and
input purification, respectively, compared with 78.4\% without
defense.

Finally, we evaluate the sensitivity of our attack to the detection confidence threshold. When the threshold varies from 0.3 to 0.7, the attack against COMO remains stable across all four datasets. Even at a strict threshold of 0.3, the Strict-ASR stays above 56.0\% on DroneVehicle. Cross-model evaluation on DroneVehicle further shows robust effectiveness across all threshold levels. The corresponding curves are provided in the Supplementary Material.

\section{Conclusion}

In this work, we presented CamoShift, an adversarial framework for visible-infrared object detection. Unlike prior cross-modal attacks that mainly rely on conspicuous patterns and perturbations confined to spatially aligned object regions, our method targets a more fundamental weakness in visible-infrared fusion. This weakness lies in the strong dependence on object-level cross-modal correspondence. Specifically, SCM generates a visually plausible patch that remains consistent with the texture and color distribution of the host object, while the OSDM shifts the infrared evidence in a scale-aware manner to break the alignment prior used by fusion networks. Together with the harmonic adversarial loss, these components jointly improve attack effectiveness and stealth, and induce severe cross-modal feature conflict during fusion.

Extensive evaluations across four benchmark datasets and three representative detector families demonstrate that CamoShift consistently outperforms existing baselines in both attack strength and visual stealthiness. The performance gains are particularly significant on fusion models that rely heavily on strict spatial alignment. This validates our core insight: disrupting spatial alignment is fundamentally more destructive than simply applying stronger, spatially aligned appearance perturbations. Transferability, camouflage-style, and component-wise ablation results further support the robustness and generality of CamoShift. Overall, our study demonstrates that spatial decoupling is a critical yet underexplored attack surface in visible-infrared detection. We hope this work inspires research into fusion mechanisms robust to spatial decoupling, defense strategies \cite{kim2022defending}, and evaluation protocols for \mbox{multi-sensor vision systems}.

{\small
\bibliographystyle{ieeenat_fullname}
\bibliography{reference}
}

\end{document}